\documentclass[conference]{IEEEtran}
\IEEEoverridecommandlockouts
\usepackage{cite}
\usepackage{amsmath,amssymb,amsfonts}
\usepackage{algorithmic}
\usepackage{graphicx}
\usepackage{textcomp}
\usepackage{xcolor}
\usepackage{multirow}
\usepackage{threeparttable}
\usepackage{url}
\usepackage{bbding}
\usepackage{array}
\usepackage{booktabs}
\usepackage{subfigure}
\def\BibTeX{{\rm B\kern-.05em{\sc i\kern-.025em b}\kern-.08em
    T\kern-.1667em\lower.7ex\hbox{E}\kern-.125emX}}
\begin{document}

\title{A Survey of Deep Learning Based Radar and Vision Fusion for 3D Object Detection in Autonomous Driving\\
{\footnotesize }
\thanks{This work was supported in part by the National Natural Science
Foundation of China (No. 61374159), Shaanxi Natural Fund (No.
2018MJ6048), Space Science and Technology Fund, the Foundation
of CETC Key Laboratory of Data Link Technology (CLDL-20182316,
CLDL20182203), and the Suzhou municipal science and technology plan project (No. SYG202351).}
}

\author{
    \IEEEauthorblockN{Di Wu$^{a}$, Feng Yang$^{a*}$, Benlian Xu$^{b}$, Pan Liao$^{a}$, Bo Liu$^{a}$}
    \IEEEauthorblockA{$^{a}$ School of Automation, Northwestern Polytechnical University, Xi'an, China}
    \IEEEauthorblockA{$^{b}$ School of Electronic and Information Engineering, Suzhou University of Science and Technology, Suzhou, China}
    \IEEEauthorblockA{\{wu\_di821\}@mail.nwpu.edu.cn, \{yangfeng\}@nwpu.edu.cn, \{xu\_benlian\}@usts.edu.cn}
}

\maketitle

\begin{abstract}
With the rapid advancement of autonomous driving technology, there is a growing need for enhanced safety and efficiency in the automatic environmental perception of vehicles during their operation. In modern vehicle setups, cameras and mmWave radar (radar), being the most extensively employed sensors, demonstrate complementary characteristics, inherently rendering them conducive to fusion and facilitating the achievement of both robust performance and cost-effectiveness. This paper focuses on a comprehensive survey of radar-vision (RV) fusion based on deep learning methods for 3D object detection in autonomous driving. We offer a comprehensive overview of each RV fusion category, specifically those employing region of interest (ROI) fusion and end-to-end fusion strategies. As the most promising fusion strategy at present, we provide a deeper classification of end-to-end fusion methods, including those 3D bounding box prediction based and BEV based approaches. Moreover, aligning with recent advancements, we delineate the latest information on 4D radar and its cutting-edge applications in autonomous vehicles (AVs). Finally, we present the possible future trends of RV fusion and summarize this paper.
\end{abstract}


\section{Introduction}
According to the latest report from the World Health Organization (WHO), approximately 1.19 million people die each year as a result of road traffic crashes\cite{Road}. Therefore, the development of reliable autonomous driving systems to assist drivers in reducing potential accidents and promptly avoiding obstacles holds significant practical significance. Autonomous driving perception refers to the ability of autonomous driving systems to perceive the surrounding environment effectively, enabling vehicles to understand and respond to various driving scenarios. This pivotal technology, essential for autonomous driving, enables intelligent vehicles to recognize, comprehend, and respond to the surrounding roads and traffic conditions, thereby significantly enhancing vehicle safety and reducing traffic accident rates. 

The Advanced Driving Assistance System (ADAS) in intelligent vehicles are usually equipped with multiple sensors including camera, radar and Lidar. Cameras can capture rich semantic information, including visual features such as the boundaries and textures of objects and backgrounds. However, they are susceptible to external factors such as weather and lighting, and their detection range is limited. Radar emits radio waves and employs the Doppler effect to precisely measure the distance and radial velocity of objects, achieving a maximum detection range of up to 250 meters\cite{wei2022mmwave}. Furthermore, radar exhibits "all-weather" capability, operating seamlessly regardless of weather conditions or time of day. In contrast to LiDAR, radar point clouds are sparse, offering computational resource savings, albeit with the trade-off of increased errors, including false positives and false negatives. As a sensor with a data format similar to that obtained by radar, LiDAR boasts superior angular resolution and denser data, thus harboring semantic information absent in radar point clouds and achieving heightened detection accuracy. Nonetheless, LiDAR is not without its drawbacks, encompassing elevated production and maintenance costs, vulnerability to environmental interference, and diminished reliability. In summary, different sensors exhibit unique strengths and weaknesses, as illustrated in Table \uppercase\expandafter{\romannumeral1}, naturally encouraging their fusion to complement each other's limitations to achieve more accurate and robust perception outcomes. 

\begin{table}
\centering
\caption{Sensor characteristics of camera, radar and LIDAR\cite{kim2020grif,kim2023craft,wei2022mmwave}.}
\label{tab:1}
\tabcolsep=0.45cm
\begin{threeparttable}
\begin{tabular}{c|ccc}
\toprule
\textbf{Sensor Type} & \textbf{Radar} & \textbf{LiDAR} & \textbf{Camera} \\
\midrule
Semantic Information & $\times$ & $\triangle$ & $\circ$  \\
Range Resolution & $\circ$ & $\circ$ & $\triangle$ \\
Angle Resolution & $\triangle$ & $\circ$ & $\circ$ \\
Weather Robustness & $\circ$ & $\triangle$ & $\times$ \\
Lighting Robustness & $\circ$ & $\triangle$ & $\times$ \\
Detection Range & $\circ$ & $\triangle$ & $\times$ \\
Velocity Information & $\circ$ & $\triangle$ & $\times$ \\
Cost & $\circ$ & $\times$ & $\circ$ \\
Maintenance & $\circ$ & $\times$ & $\circ$ \\
Reliability & $\circ$ & $\times$ & $\circ$ \\
\bottomrule
\end{tabular}
 \begin{tablenotes}
        \footnotesize
        \item $\circ$: Good, $\triangle$: Normal, $\times$: Bad
      \end{tablenotes}
  \end{threeparttable}
\end{table}

Currently, although the cost of LiDAR has slightly decreased, its application in intelligent vehicles is still constrained by factors such as extreme weather, poor lighting conditions, and interference from similar frequencies\cite{Pros}. Meanwhile, due to the advantages of low cost, maintainability, and high reliability of cameras and radar, they dominate the sensor setup in Level 2 to Level 3 autonomous vehicles (as classified by the J3016 standard released by the Society of Automotive Engineers (SAE) in 2014 \cite{sae2013definitions}), making radar-vision (RV) fusion perception a primary research focus. Recently, significant progress has been made in the next generation of automotive radar, which not only provides distance, azimuth, and radial velocity but also offers elevation data, referred to as 4D radar. 4D radar tends to provide denser point clouds\cite{meyer2019automotive}, complemented by precise height information, making its data similar to LiDAR but with additional velocity details. These characteristics contribute to the application of 4D radar in autonomous driving perception, helping to overcome limitations associated with traditional mmWave radar.

Several surveys have been conducted on RV fusion perception. Wang et al. \cite{wang2019multi} cover strategies for radar-vision data fusion but lack a review of the currently most prevalent fusion techniques based on deep learning. Wei et al.\cite{wei2022mmwave} provide a detailed review of RV fusion object detection, while Tang et al.\cite{tang2021road} not only review object detection but also cover relevant content on object tracking. However, both lack a systematic overview of the latest 3D object detection. Singh et al.\cite{singh2023vision} focus on Bird's-Eye-View (BEV) detection, which is an advanced solution in 3D object detection in recent years. However, it includes a significant amount of 2D object detection methods and lacks a comprehensive summary of recent advancements in 3D object detection. In this paper, with the background of autonomous driving perception, we provide a more comprehensive review of 3D object detection techniques based on modern deep learning for RV fusion. The main contributions can be summarized as follows:

\begin{itemize}
\item A detailed discussion of the latest deep learning-based RV fusion methods for 3D object detection in autonomous driving scenarios.

\item Differing from traditional fusion strategies categorized based on the stages of integrating information from different sensors, this paper classifies RV fusion methods into two categories based on the characteristics of deep learning frameworks: region of interest (ROI) based and end-to-end. 

\item In alignment with the latest technological developments, the paper compiles the most recent research advancements regarding the application of 4D mmWave radar in AVs. 
\end{itemize}

As shown in Fig. 1, the rest of this article is organized as follows. Section \uppercase\expandafter{\romannumeral2} introduce the tasks of object detection, RV fusion datasets and the others. Section \uppercase\expandafter{\romannumeral3} reviews RV fusion schems for object detection, including ROI-based fusion and end-to-end fusion. Section \uppercase\expandafter{\romannumeral4} provides the latest applications of 4D radar in AVs. Section \uppercase\expandafter{\romannumeral5} analyzes the future trends of RV fusion in AVs. Section \uppercase\expandafter{\romannumeral6} concludes this article. 

\begin{figure}
    \centering
    \includegraphics[width=2.8 in]{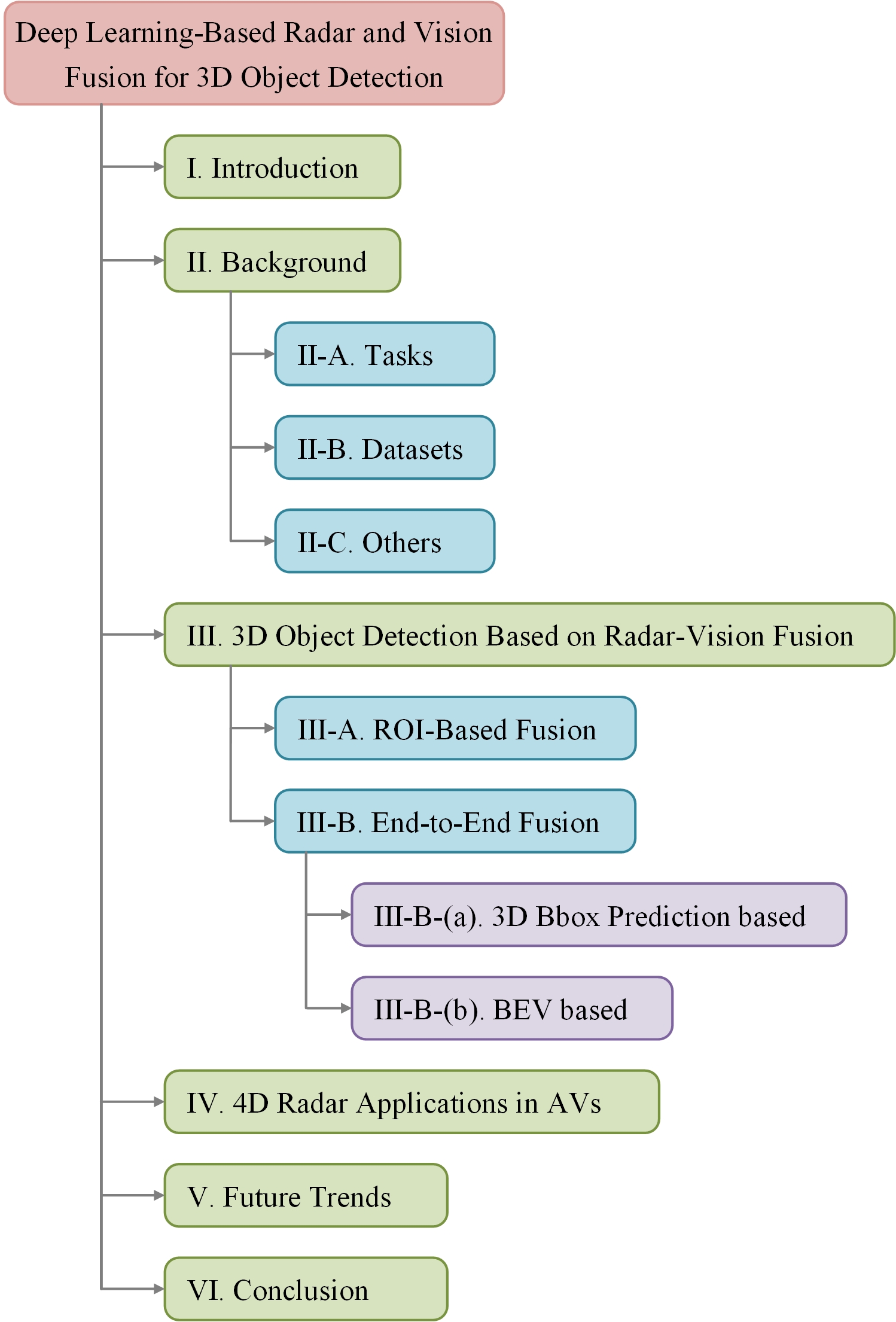}
    \caption{The organization of this paper.}
    \label{fig:1}
\end{figure}

\section{Background}

\subsection{Tasks}
Object detection, a core challenge in computer vision, involves locating objects of predefined categories in an image, regressing their bounding boxes, determining their classes, and, in the context of autonomous driving, estimating their velocities to enhance driving safety. Object detection can be unified into the following paradigm\cite{mao20233d}:

\begin{equation}
\mathcal{B}=f_{det}(\mathcal{I}_{sensor}),
\end{equation}
where $\mathcal{B}=\{B_1,...,B_N\}$ is a set of $N$ objects in a scene, $f_{det}$ is a object detection model, and $\mathcal{I}_{sensor}$ is one or more sensory inputs. The current on-road object detection tasks can be divided into two types: 2D object detection and 3D object detection. The object representations $B_{i}$ for these two tasks are different  which determines the information available for subsequent prediction and planning steps. In most cases, the representations of a 2D object and a 3D object are, respectively, a rectangular bounding box in an input image and a cuboid in a three-dimensional coordinate system as shown in Fig. 2, denoted as $B_i^{2D}$ and $B_i^{3D}$, represented as

\begin{equation}
B_i^{2D}=[x_r,y_r,w_r,h_r,class],
\end{equation}
\begin{equation}
B_i^{3D}=[x_c,y_c,z_c,l_c,w_c,h_c,\theta,class],
\end{equation}
where $(x_r,y_r)$ and $(w_r,h_r)$ are the 2D center coordinate and the shape parameters of a rectangular bounding box respectively, $(x_c,y_c,z_c)$ and $(l_c,w_c,h_c)$ are the 3D center coordinate and the shape parameters of a cuboid respectively, $\theta$ is the heading angle of a cuboid and $class$ denotes the category of the object. In some certain scenarios, particularly in on-road object detection, the velocity components of an object $(v_x,v_y)$ along the ground plane are also considered as supplementary parameters. 

\begin{figure}
    \centering
    \includegraphics[width=3.5 in]{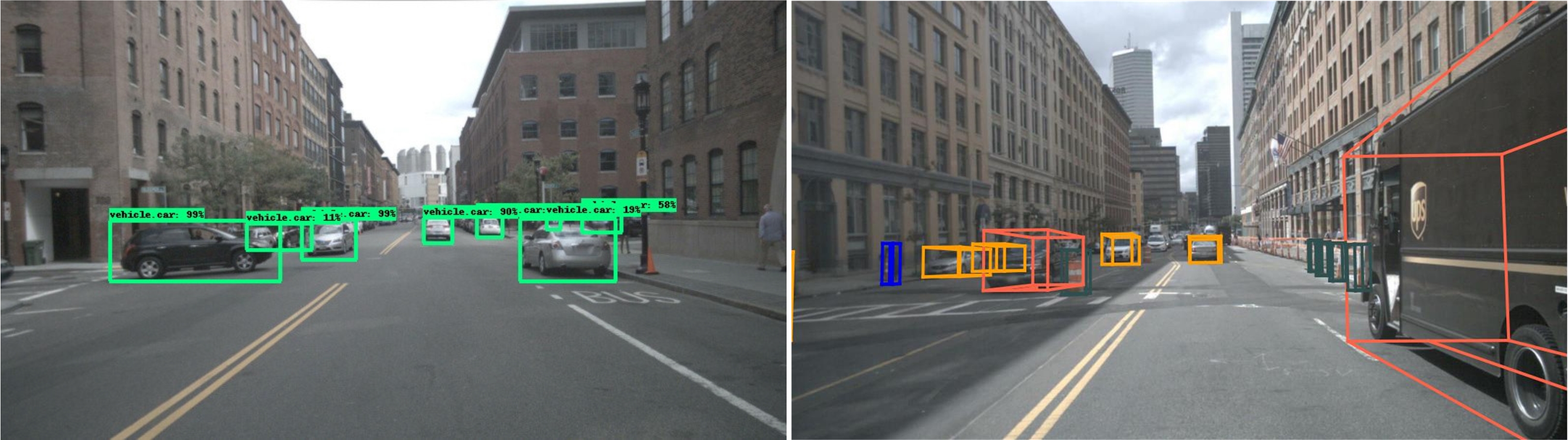}
    \caption{2D and 3D object detection results. Left: 2D object detection\cite{li2020feature}. Right: 3D object detection\cite{li2022bevformer}.}
    \label{fig:2}
\end{figure}

\subsection{Datasets}
The dataset serves as the research foundation for deep learning-based object detection. Table \uppercase\expandafter{\romannumeral2} summarizes some widely used public datasets with both camera and radar setups for autonomous driving scenarios. Additionally, it encompasses the latest datasets that include 4D radar point clouds. The term "Radar" in the table refers to traditional mmWave radar, distinguished from "4D Radar" that includes height information. The nuScenes\cite{caesar2020nuscenes} dataset is one of the most commonly used and sensor-rich autonomous driving datasets established by nuTonomy. The vehicle used to collect this dataset is equipped with six surrounded cameras, five surrounded radars and one lidar covering the 360-degree field of view. It provides annotations for 23 classes of 3D bounding boxes in road scenes. Apart from the nuScenes dataset, all other datasets have cameras with only a forward-facing view. The RadarScene\cite{schumann2021radarscenes} dataset has a large number of annotated frames but lacks LiDAR data. The Astyx\cite{meyer2019automotive} dataset is the first to utilize a new generation of 4D radar. Unfortunately, it has a limited number of labeled frames, and its class distribution is imbalanced, making it unsuitable for most data-driven deep learning object detection research. The VoD\cite{palffy2022multi} dataset and the TJ4DRadSet\cite{zheng2022tj4dradset} dataset are two novel autonomous driving datasets featuring complex urban road scenes, containing numerous frames of synchronized and calibrated camera, LiDAR and 4D radar data. K-Radar\cite{paek2022k} is the latest large-scale object detection dataset that contains 35K frames of 4D Radar tensor data. 
However, there is still a lack of multimodal datasets similar to nuScenes that cover a $360^{\circ}$ surrounded view for each type of sensor, which is crucial for research on surround perception based on RV fusion. Additionally, due to the development of data-driven large model techniques, large-scale and scene-rich datasets are still in demand.

\begin{table*}
\centering
\caption{Current driving datasets with camera and radar setups.}
\label{tab:2}
\tabcolsep=0.28cm
\begin{threeparttable}
\begin{tabular}{cccccccc}
\toprule[1pt]
\specialrule{0em}{1pt}{1pt}
\textbf{Dataset} & \textbf{Year} & \textbf{Camera} & \textbf{Radar} & \textbf{LIDAR} & \textbf{4D Radar} & \textbf{Labeled Frames} & \textbf{3D Annotations}\\
\specialrule{0em}{1pt}{1pt}
\midrule[0.5pt]
\specialrule{0em}{1pt}{1pt}
STF\cite{bijelic2020seeing} & 2020 & stereo, front & 1, front & 2 & - & 13.5k & \CheckmarkBold\\
\specialrule{0em}{1pt}{1pt}
nuScenes\cite{caesar2020nuscenes} & 2019 & 6$\times$mono, surround & 5, surround & 1 & - & 400k & \CheckmarkBold\\
\specialrule{0em}{1pt}{1pt}
RADIATE\cite{sheeny2021radiate} & 2020 & stereo, front & 1, surround, no velocity & 1 & - & 44k & \XSolidBrush\\
\specialrule{0em}{1pt}{1pt}
RadarScenes\cite{schumann2021radarscenes} & 2021 & mono, front & 4, front/side & \XSolidBrush & - & 832k & \XSolidBrush\\
\specialrule{0em}{1pt}{1pt}
\midrule[0.5pt]
\specialrule{0em}{1pt}{1pt}
Astyx\cite{meyer2019automotive} & 2019 & mono, front & - & 1 & 1, front & 546 & \CheckmarkBold\\
\specialrule{0em}{1pt}{1pt}
VoD\cite{palffy2022multi} & 2021 & stereo, front & - & 1 & 1, front & 8693 & \CheckmarkBold\\
\specialrule{0em}{1pt}{1pt}
TJ4DRadSet\cite{zheng2022tj4dradset} & 2022 & mono, front & - & 1 & 1, front & 7757 & \CheckmarkBold\\
\specialrule{0em}{1pt}{1pt}
K-Radar\cite{paek2022k} & 2023 & 4$\times$stereo, surround & - & 2 & 1, front & 35k & \CheckmarkBold\\
\specialrule{0em}{1pt}{1pt}
\bottomrule[1pt]
\end{tabular}
  \end{threeparttable}
\end{table*}

\subsection{Others}
The evaluation metrics and input-data formats have been extensively discussed in detail in \cite{singh2023vision}, and this paper will not repeat them. It is worth noting that in Section \uppercase\expandafter{\romannumeral3}-B of \cite{singh2023vision}, the general form of radar point cloud $P=(x,y,z,v_x,v_y)$ is presented, where the height information $z$ provided by traditional mmWave radar is inaccurate, but 4D radar can offer precise $z$ values.

\section{3D Object Detection Based on Radar-Vision Fusion}
Fusing data from different sensors can fully leverage their respective perceptual advantages, leading to more robust performance in object detection compared to relying on a single type of sensor. This section provides a comprehensive overview of the deep learning-based 3D object detection methods with RV fusion frameworks for autonomous vehicles (AVs). Fusion methods include two types: region of interest (ROI) and end-to-end based. From a reliability standpoint, in the event of a type of sensor failure, the ROI-based fusion strategy can rely solely on proposals from the other type of sensor to detect objects. However, in most scenarios, the end-to-end fusion strategy can more comprehensively integrate features from both sensor data.

\subsection{ROI-Based Fusion}\label{AA}
The ROI-based fusion initially uses input information from one type of sensor to generate preliminary regions of interest (ROIs). Subsequently, features corresponding to these ROIs are further extracted from the data of another sensor or the fused multimodal data for the classification and refinement. Due to the simplicity of the 3D to 2D transformation, which only requires a transformation matrix without additional information, most ROI-based fusions adopt the workflow of extracting ROIs from radar point clouds and subsequently projecting them onto images for feature extraction. This method, exemplified by the works cited in \cite{nabati2019rrpn, ji2010incremental, wang2014bionic, yadav2020radar+}, is referred to as radar-generated ROIs, depicted in Fig. 3(a). Another strategy detects objects on the image first, then crops image patches based on the bounding boxes, and finally complements velocity and other information using radar points, as done in \cite{wang2018vehicle,jha2019object}, termed vision-generated ROIs and illustrated in Fig. 3(b).

\begin{figure}
    \centering
    \subfigure[]{\includegraphics[width=3.5 in]{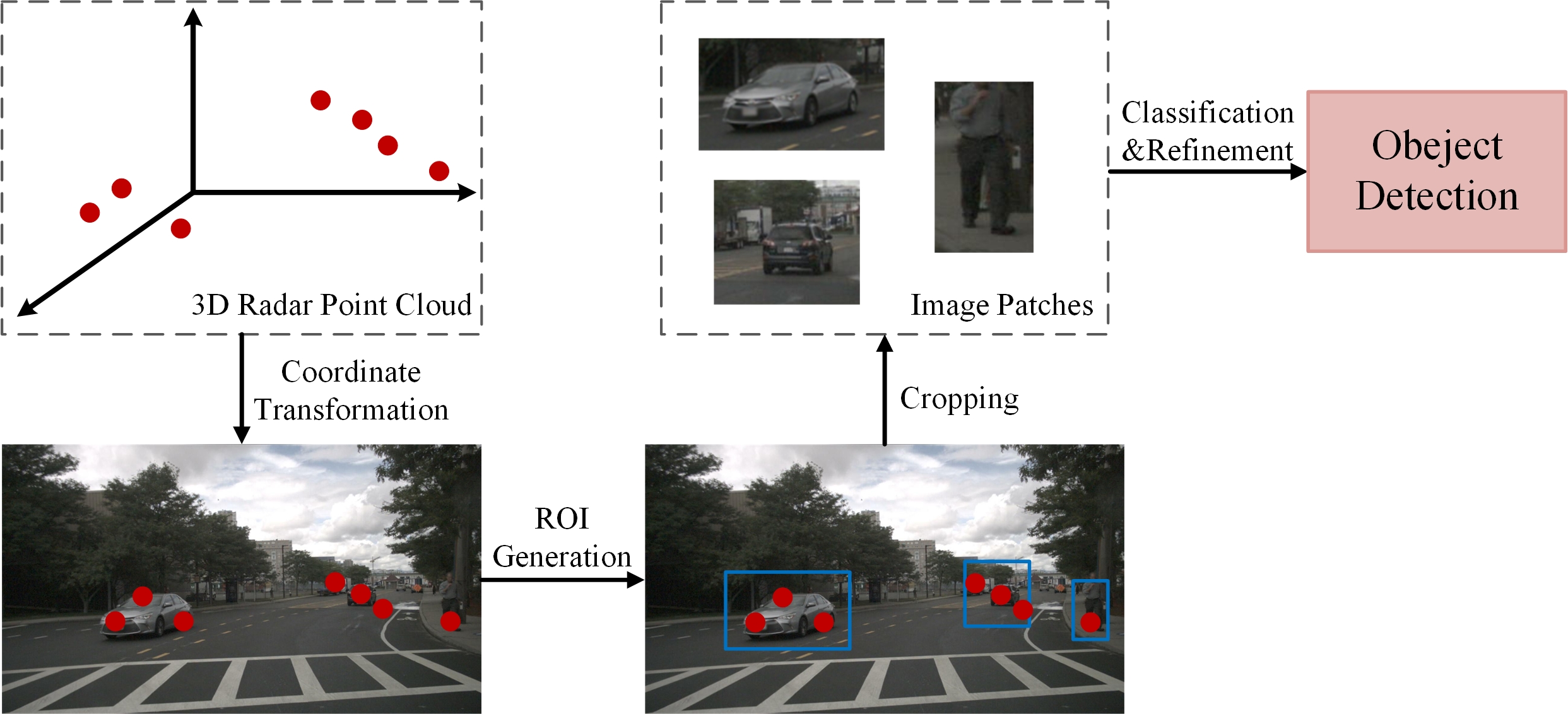}
    \label{fig:a}}
    \subfigure[]{\includegraphics[width=3.5 in]{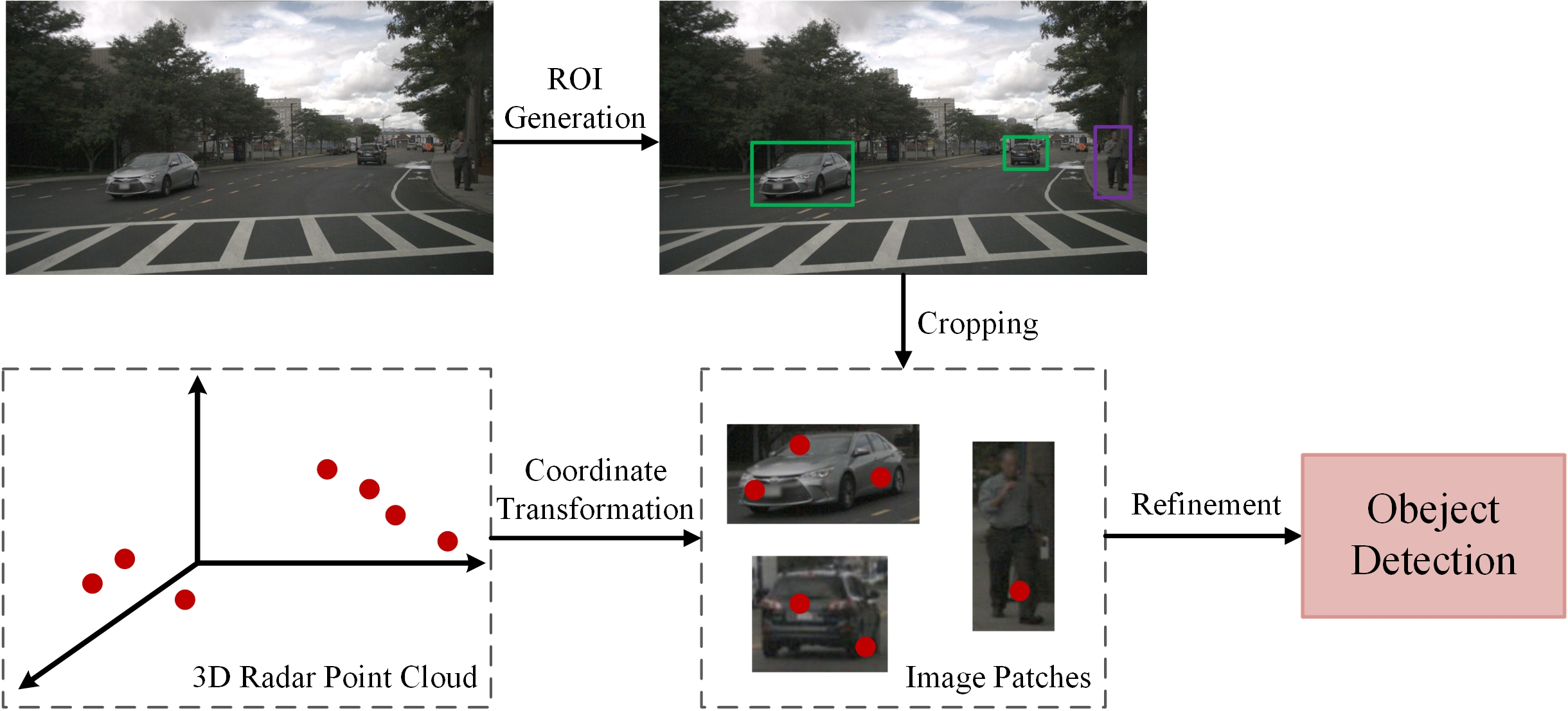}
    \label{fig:b}}
    \caption{The ROI-based RV fusion frameworks ((a): Radar-generated ROIs; (b): Vision-generated ROIs).}
\end{figure}

The radar-generated ROIs strategy, which directly projects radar points onto the 2D image plane, can flatten the depth dimension of the point cloud, potentially leading to the loss of some 3D spatial information originally captured by the sensor \cite{kurniawan2023clusterfusion}. Therefore, such methods are more suitable for 2D object detection where depth value is not required. For 3D object detection, an increasing amount of work focuses on the vision-generated ROIs scheme. CenterFusion\cite{nabati2021centerfusion} introduces a pillar expansion step to address the inaccurate height information of radar points. Then, it associates radar detections with the corresponding object proposals using a frustum-based method, supplementing image features and regressing object properties. ClusterFusion\cite{kurniawan2023clusterfusion} also generates preliminary 3D object detections from the input image in first stage and then leverages the radar points to refine the preliminary detections’ velocity, depth, orientation, and attribute prediction. However, these approaches tend to rely on the preliminary locations of the objects in the world coordinate system from images, which inherently entail considerable uncertainty due to the lack of depth information. Consequently, sampling radar points based on these imprecise positions and directly discarding unassociated radar points negatively affects detection accuracy. CRAFT\cite{kim2023craft} associates 3D proposals generated from images with radar points in the polar coordinate system through soft association. Subsequently, through consecutive cross-attention-based feature fusion layers, it adaptively exchanges spatial-contextual information between the camera and radar to address incorrect associations, significantly improving detection accuracy. The above methods all suggest that maximizing the number of correct associations between 3D proposals and radar points appears to be a key step in ROI-based RV fusion framework. However, due to the sparsity of radar points and their lack of height information, optimizing this crucial step has limited impact on enhancing model performance. Benefiting from the advancements in modern radar technology, the resolution of radar sensor has also gradually improved, resulting in denser radar point clouds than before. This progress makes it possible to adapt existing mature LiDAR-based architectures\cite{qi2017pointnet,qi2017pointnet++,zhou2018voxelnet,lang2019pointpillars} to handle radar data \cite{danzer20192d, dreher2020radar}. Nonetheless, compared to LiDAR point clouds, radar point clouds remain highly sparse and lack sufficient semantic information. Therefore, ROI-based fusion frameworks are limited in performance. Regardless of which modality generates the regions of interest, significant detection errors can occur due to the lack of depth or height information, directly impacting the final object detection results. However, these fusion frameworks also offer some benefits, as they can partially reduce the search scope for object detection, thereby saving computational resources.

\subsection{End-to-End Fusion}
The end-to-end fusion strategy processes data from both the camera and radar simultaneously. By integrating features from both modalities within a unified framework and leveraging their complementary advantages, the perception performance becomes more robust. This type of approach is currently one of the most prominent fusion pipelines. The fundamental framework of end-to-end RV fusion is illustrated in Fig. 4. We further categorize this approach into two sections: 3D bounding box prediction based and Bird's-Eye-View (BEV) based.

\begin{figure}
    \centering
    \includegraphics[width=3.5 in]{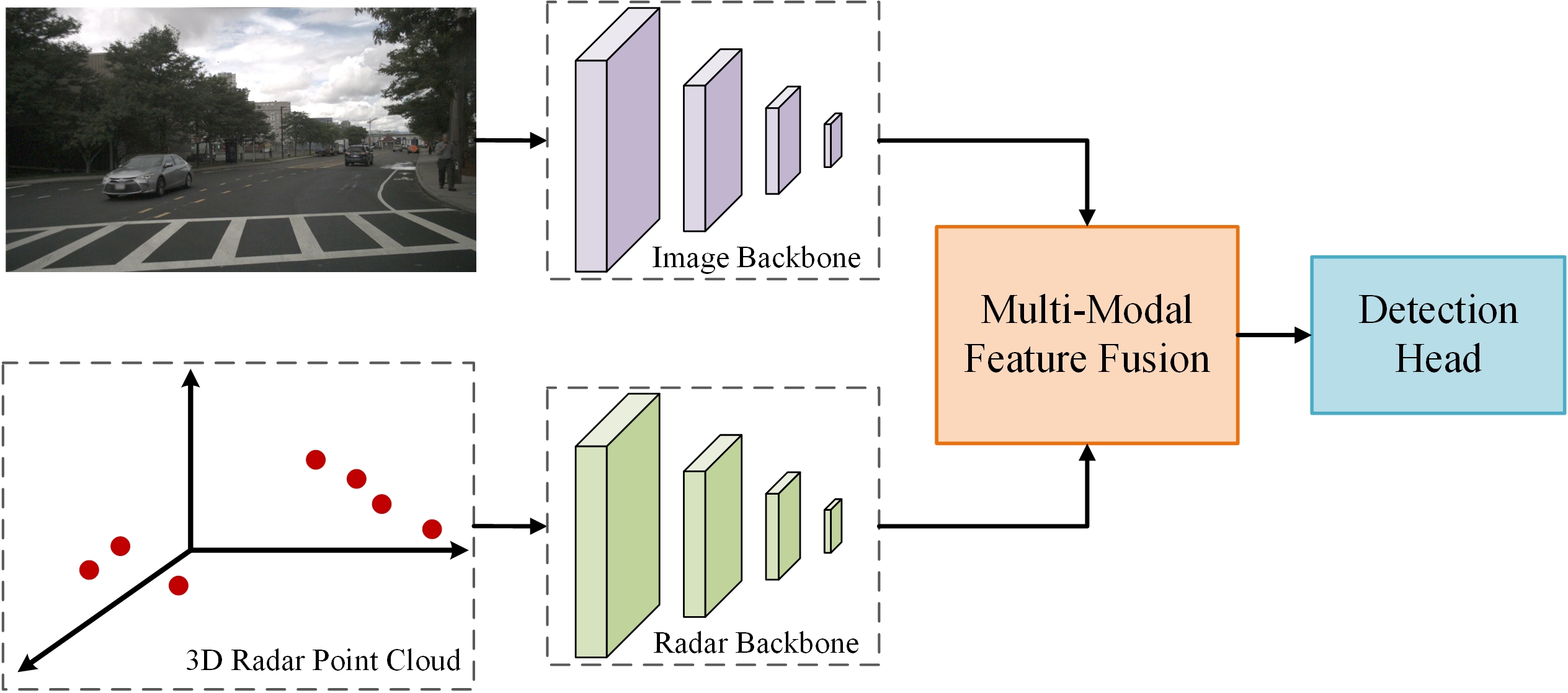}
    \caption{The general end-to-end RV fusion framework.}
    \label{fig:4}
\end{figure}

\paragraph{3D Bbox Prediction based}
Thanks to the development of mature modern 2D detection techniques, many RV fusion frameworks directly incorporate separate input branches for processing radar data into advanced 2D detection networks and fuse radar features with image features in the middle of the network. For example, an additional branch was added to the SSD detection framework for radar input data in \cite{chadwick2019distant}, and \cite{li2020feature} and \cite{john2019rvnet} extend the input channels based on the YOLO series networks to simultaneously extract image and radar features. CRF-Net\cite{nobis2019deep} employs VGG16 as the model backbone, utilizes auxiliary branches to extract radar features at different levels. Undoubtedly, the most straightforward to accomplish 3D object detection task is to transplant existing mature 2D computer vision framework and graft on a 3D detection head, which is indeed the case in practice. In the early stages of the development of 3D object detection, researchers focused on utilizing various convolutional neural networks (CNNs) to simultaneously extract features from both images and radar data. However, unlike 2D object detection, 3D object detection models need to regress 3D bounding boxes in three-dimensional space, including length, width, and height information. In \cite{meyer2019deep}, a 3D region proposal network is used to generate proposals based on camera images and radar images. GRIF Net \cite{kim2020grif} predefines 3D anchor boxes of different heights and sizes and projects them onto both the camera perspective and radar BEV views. Then, it utilizes a 3D region proposal network (RPN) to generate 3D proposals. However, these two methods do not effectively leverage the correlation between camera features and radar features. They merely learn the parameters of 3D bboxes directly from the features of the two sensors and the concatenated fusion features. Recently, the introduction of attention mechanisms has brought further improvements to the performance of computer vision models. To resolve the ambiguity in the geometric correspondences between radar and camera features, the authors propose a ray-constrained cross-attention mechanism to better utilize radar range measurements for better camera depth predictions in \cite{hwang2022cramnet}. SparseFusion3D \cite{yu2023sparsefusion3d} builds upon the architecture of DETR3D \cite{wang2022detr3d} by using radar points to initialize object queries, and the 3D reference points decoded from object queries are projected into the image space to extract image features. Overall, 3D bbox prediction based methods draw inspiration from many excellent ideas in 2D object detection networks. However, predicting 3D bboxes entails estimating more parameters intricately linked to three-dimensional space, often demanding increased computational resources and more complex algorithms.

\paragraph{BEV based}
Recently, due to its ability to provide a panoramic and unobstructed perception perspective, Bird's-Eye-View (BEV) perception schemes have gradually taken a dominant position in 3D object detection. BEV-based methods simplify object detection to operate on two-dimensional images from a top-down perspective, enabling the utilization of rich techniques and algorithms in the field of computer vision while also improving computational efficiency. Many works consider leveraging radar detections with robust depth sensing to assist in the transformation of image features from perspective view to BEV. In \cite{kim2023crn}, the authors lift the image features to three-dimensional space using the predicted depth distribution and concatenate them with the image BEV features along the height channel guided by radar depth priors utilizing the radar BEV occupancy. Then, through a deformable cross-attention module, they adaptively fuse the image BEV features and radar BEV features to handle noisy and ambiguous radar points. This work leverages the depth sensing advantage of radar to complement a monocular depth estimation network. However, it relies on two parallel and independent view transformations, which inevitably leads to spatial misalignment between BEV features from the two modalities. RCM-Fusin\cite{kim2023rcm} adopts BEVFormer \cite{li2022bevformer} as the baseline and creates an optimized BEV query containing radar position information extracting from the radar BEV feature map via a deformable self-attention mechanism \cite{zhu2020deformable}, thus integrating features from both modalities to accomplish implicit view transformation. In \cite{wolters2024unleashing}, the authors associate the pillar features with sparse depth encoding from radar point cloud with the corresponding depth-missing image columns using cross-attention to generate unified geometry-aware features in the perspective view. They then refine the initial BEV queries using radar-weighted depth consistency computed from radar BEV features, tackling the problem of misaligned or associated features. The key challenge in these approaches lies in how to leverage the depth information from radar points to improve the perception of depth in perspective view features and how to handle the spatial misalignment between image and radar BEV features. HVDetFusion \cite{lei2023hvdetfusion} is a two-stage detection framework. In the first stage, it transforms image features from 2D space to 3D space using estimated depth. It then utilizes the first detection head to obtain preliminary detection results, which are used as prior information to refine false positives in the initial radar data. Subsequently, it integrates radar detection with image detection and utilizes the second detection head to output fused detection results. This is currently the state-of-the-art method for radar-camera fused 3D object detection in the nuScenes leaderboard.

\section{4D Radar Applications in AVs}
With the advancement of radar technology, 4D radar has addressed the deficiency of traditional radar in lacking height information, which has attracted researchers' attention, leading to a gradual exploration of how to apply it in AVs. In \cite{palffy2022multi}, the authors apply PointPillars, previously used for LiDAR 3D data, to 4D radar data for multi-class road user detection. MVFAN\cite{yan2023mvfan} is an end-to-end and single-stage framework for 3D object detection utilizing the Radar Feature Assisted backbone to fully exploit the valuable 4D radar data. RCFusion\cite{zheng2023rcfusion} achieves camera and 4D radar features fusion under a unified BEV space, which introduces a radar PillarNet to generate radar pseudo images. A fusion module named IAM is then used to fuse both BEV feature types adaptively. Apart from object detection, there are also efforts to utilize 4D radar for other autonomous driving tasks. CenterRadarNet\cite{cheng2023centerradarnet} is a joint 3D object detection and tracking framework using 4D radar, consisting of a single-stage 3D object detector and an online re-identification (re-ID) tracker. 4DRVO-Net\cite{zhuoins20234drvo} is a method for 4D radar visual odometry that integrates camera and 4D radar information. It involves the design of an adaptive 4D radar$-$camera fusion module (A-RCFM) that automatically selects image features based on 4D radar point features. The method proposed in \cite{li2024radarcam} combines image and 4D radar point cloud fusion for metric dense depth estimation. In summary, 4D radar point clouds, as a robust sensor data with higher density than traditional 3D radar and additional Doppler information compared to LiDAR, deserve further exploration. However, similar to 3D radar, 4D radar point clouds are still relatively sparse. Establishing accurate associations and feature interactions between 4D radar point clouds and images remains a significant challenge. 

\section{Future Trends}
Through the review and analysis in this article, we believe that the RV fusion perception in the context of autonomous driving has the following development trends:

\paragraph{End-to-End Autonomous Driving} 
End-to-end autonomous driving directly takes raw sensor data as input and integrates tasks such as perception, path planning, control, and decision-making into a single neural network for learning. It directly outputs the operational instructions required to control the vehicle's behavior, without the need for manually designing complex intermediate representations or processing steps. This technology eliminates the complex modular structure in traditional autonomous driving systems, simplifying the system design and implementation process. Moreover, because it can automatically discover complex patterns and features in sensor data, it better understands the environment and makes more accurate decisions.

\paragraph{Application of 4D Radar} 
With the advancement of 4D mmWave radar technology, it is progressing towards higher resolution and longer detection ranges. Due to its cost advantages, it may replace traditional radar and LIDAR in some mass-produced intelligent vehicles in the future. This brings the need for a more concise and efficient fusion solution between 4D radar and vision. The challenges and trends in research include how to deeply integrate two heterogeneous multimodal data sources and how to improve the real-time performance of perception systems while maintaining accuracy. 

\paragraph{Collaborative Perception} 
Collaborative perception refers to the process in which multiple AVs exchange information and cooperate to collectively perceive the surrounding environment and make decisions. This innovative perception concept empowers vehicles on the road to achieve real-time and comprehensive environmental awareness. It not only improves the safety and reliability of autonomous driving systems but also optimizes the efficiency of the overall transportation system, aligning seamlessly with the requirements of smart transportation development.

\section{Conclusion}
Perception, as one of the three key modules in autonomous driving systems, plays a crucial role in processing information from multiple sensors and extracting relevant environmental data needed by the other two modules: control and decision-making. As the most common low-cost sensors in mass-produced vehicles, cameras and radars possess rich semantic information and all-weather operational characteristics, and their complementary advantages can achieve a relatively ideal perception performance. In this paper, we first analyze the strengths and weaknesses of several types of sensors, and then introduce existing publicly available datasets that incorporate both radar and camera, including the latest 4D radar datasets. We then provide a detailed review of the current status of RV fusion-based 3D object detections. The 3D object detection techniques based on deep learning are categorized into two strategies: ROI-based and end-to-end. To align with the latest technologies, we present the recent applications of 4D radar in the autonomous driving industry. Finally, we analyze possible trends in the development of autonomous driving RV fusion perception for the reference of readers.

\vspace{12pt}

\end{document}